\documentclass{article} 
\usepackage[final]{colm2026_conference}

\usepackage{microtype}
\usepackage{hyperref}
\usepackage{url}
\usepackage{booktabs}
\usepackage{amsmath}
\usepackage{amssymb}
\usepackage{xcolor}
\usepackage{tikz}
\usetikzlibrary{arrows.meta,positioning,shapes.geometric,calc,fit,backgrounds,shadows.blur,decorations.pathreplacing,shapes.misc}
\usepackage{algorithm}
\usepackage{algpseudocode}
\usepackage{placeins}

\usepackage{lineno}

\definecolor{darkblue}{rgb}{0, 0, 0.5}
\hypersetup{colorlinks=true, citecolor=darkblue, linkcolor=darkblue, urlcolor=darkblue}

\definecolor{easyc}{RGB}{46,160,67}    
\definecolor{medc}{RGB}{221,145,26}    
\definecolor{hardc}{RGB}{205,58,58}    
\definecolor{calibc}{RGB}{58,124,190}  
\definecolor{sampc}{RGB}{124,92,186}   
\definecolor{outc}{RGB}{38,150,148}    

\title{THRIFTY: Cost-Constrained Reasoning-Data Synthesis \\ for Small Language Models}

\author{Aasheesh Singh \\
Phi Labs, Quantiphi \\
Toronto, ON, Canada
\And
Vishal Vaddina \\
Phi Labs, Quantiphi \\
Toronto, ON, Canada
}

\begin{document}

\ifcolmsubmission
\linenumbers
\fi

\maketitle

\begin{abstract}
Small language models (SLMs) are increasingly attractive for efficient, specialized
deployment, and are typically trained on reasoning trajectories distilled from a larger
teacher LLM. Generating this synthetic data is a major cost of the pipeline, especially
when the teacher is a frontier model. Existing collection methods, however, are largely
cost-unaware, or at best spread a fixed sampling budget uniformly across queries. Such allocation ignores the wide variation in problem difficulty and the strategic over- or under-sampling it invites. Thus, we propose THRIFTY, a Thompson-sampling bandit controller that spends a fixed generation-token budget to collect diverse correct reasoning trajectories, casting the task as a bandits-with-knapsacks problem with a novel diversity-aware objective. It tracks each problem's success probability with an online Beta posterior and spends each token where it expects the most new distinct correct solutions. Because repeated solutions add progressively less value, THRIFTY adaptively stops resampling saturated problems, concentrates budget on under-covered ones, and retires the unsolvable at the budget's opportunity cost. On math (AIME, MATH-500) and code (LiveCodeBench) benchmarks, THRIFTY gathers up to $22\%$ more diverse correct trajectories per generation token than both the standard uniform-sampling recipe and the strongest adaptive baseline, with the largest gains in the scarce-budget and hard-problem regimes.
\end{abstract}

\section{Introduction}

Small language models (SLMs) are increasingly adopted for efficient, specialized LLM deployment, and a dominant recipe for giving them strong reasoning ability is to train them on chains of thought produced by a larger teacher model. There are two costs associated with this process: the data generation cost, and the student model training cost. While a lot of research focuses on creating optimal pipelines for training the model, creating cost efficient data generation pipelines is often understudied.  Standard rejection-sampling based pipelines draw a fixed number of samples from every problem, independent of how hard it is or how much a further sample would add. Such uniform allocation is doubly
wasteful: easy problems keep being sampled long after a correct solution has been found and merely accumulate near-duplicate traces, while unsolvable problems are sampled just as
heavily and return nothing. The line of work that does adapt sampling has largely optimized
\emph{test-time} accuracy, choosing how much compute to spend answering a query at inference, rather than the distinct problem of assembling a diverse and correct training set
offline under a fixed cost.

 Toward this end, we introduce THRIFTY, a two-stage bandit controller that spends a fixed generation-token budget to
  collect the most valuable set of \emph{distinct} correct reasoning traces the budget allows. THRIFTY treats
each problem as an arm of a multi-armed bandit and casts the task as a bandits-with-knapsacks
problem \citep{badanidiyuru2018bandits}: it maintains a Beta posterior over each problem's
per-sample success probability and, at each step, spends the next token on the problem with
the highest expected marginal value per token. Because that value is diversity-gated and
diminishes gracefully if enough high quality distinct trajectories are obtained, the budget is adaptively concentrated on hard but solvable problems, while retiring hopeless or unsolvable problems outside of the policy model's capability.

\paragraph{Contributions.} We summarize our main contributions below:
\begin{enumerate}
    \item We formulate the collection of correct teacher reasoning trajectories under a fixed
    generation-token budget as a bandits-with-knapsacks problem, casting each problem as a
    bandit arm and allocating every token to the arm with the highest expected marginal
    value. Unlike prior adaptive-sampling work (\cite{zuo2025strategic}) aimed at test-time accuracy, our objective is
    offline training-data synthesis: a diversity-gated, diminishing returns value function over \emph{distinct}
    correct solutions that rewards coverage over repetition, and drives the controller to oversample hard-but-solvable problems and
    retire unsolvable ones at the budget's opportunity cost.

    \item We introduce a data-driven calibration stage that removes the hand-tuned knobs
    endemic to reasoning-data generation. Rather than presetting the generation length,
    sampling configuration, redundancy threshold, query difficulty priors to
    dataset-agnostic defaults, THRIFTY spends a small probe budget to estimate each of them
    directly from the target data. We fold this probe cost into the budget we report and show
    that it is amortized: the settings it recovers make the sampling stage efficient enough to
    come out ahead of uniform allocation even after paying for calibration.

    \item We conduct extensive experiments on math (AIME, MATH-500) and code (LiveCodeBench)
    benchmarks with a Qwen2.5 32B teacher. Our analyses show that THRIFTY
    collects more distinct correct trajectories per token than uniform allocation, with the
    largest gains in scarce-budget and hard-problem regimes.
\end{enumerate}

\section{Related Work}
\label{sec:related}

\paragraph{Synthetic reasoning data and self-training.} A now-standard way to build reasoning
training data is to sample solutions from a capable model, keep the correct ones, and
fine-tune on them, as in STaR \citep{zelikman2022star}, rejection-sampling fine-tuning
\citep{yuan2023scaling}, ReST \citep{gulcehre2023reinforced}, ReST-EM \citep{singh2024beyond},
and reward-ranked variants \citep{dong2023raft}. These pipelines establish the
sample-filter-train loop we accelerate, but they sample uniformly, drawing the same number of
trajectories from every problem with no notion of a generation budget. RFT
\citep{yuan2023scaling} is our closest value anchor: it shows accuracy scales with the number
of \emph{distinct} correct reasoning paths per problem, which motivates our diversity-gated
objective. We differ by collecting those distinct solutions online under an explicit token
budget rather than by uniform oversampling and post-hoc deduplication. A complementary line
chooses \emph{which} model to sample from, showing weaker but cheaper teachers can be
compute-optimal \citep{bansal2024smaller}; we instead fix the teacher and optimize how its
budget is spent across problems.

\paragraph{Adaptive test-time compute.} A growing body of work allocates inference compute by
problem difficulty \citep{snell2024scaling,damani2025learning}, motivated by the steep
per-problem heterogeneity of repeated sampling \citep{brown2024large} and building on
aggregation methods such as self-consistency \citep{wang2023selfconsistency}. Closest to us,
\citet{zuo2025strategic} cast test-time compute allocation as a multi-armed bandit with one
arm per query and stop sampling provably unsolvable queries. We share this bandit machinery
but target a different objective: rather than maximizing the number of queries answered at
inference, we assemble a diverse, correct \emph{training} set offline, where the currency is
generation tokens, the value is concave in distinct solutions, and both solved-enough and
unsolvable problems are retired.

\paragraph{Difficulty-aware and curriculum sampling.} Curriculum and learning-progress
methods steer training effort toward problems of intermediate, informative difficulty
\citep{matiisen2020teacher,jiang2021prioritized}, and \citet{graves2017automated} in
particular treat each data category as a bandit arm with a learning-progress reward, the
closest precedent to a bandit that allocates a training budget. In the RL-for-reasoning
setting, \citet{bae2025online} show expected improvement is lower-bounded by the variance of
task success probabilities, which peaks at a pass rate near one half, while DAPO
\citep{yu2025dapo} and DeepSeek-R1 \citep{deepseekai2025r1} discard prompt groups that are
entirely correct or entirely wrong. These operate at the level of RL batches or fixed
heuristics; our controller estimates each problem's difficulty online with a Beta posterior
and turns those estimates into per-token allocation and evidence-based retirement decisions.

\paragraph{Diversity-aware data selection.} Measuring and promoting diversity is well
studied, from Self-BLEU \citep{zhu2018texygen} and distinct-$n$ \citep{li2016diversity} to
submodular and determinantal formulations \citep{kulesza2012determinantal}, semantic
deduplication \citep{abbas2023semdedup}, and quality-weighted diversity selection
\citep{bukharin2024data}. This work almost always selects a diverse subset from a fixed pool
offline, or deduplicates after generation. We instead make diversity a first-class, online
objective: a concave value over distinct correct solutions, evaluated per problem inside the
sampling loop, so the controller stops spending on a problem the moment further samples cease
to add new solutions.

\paragraph{Bandit and active-learning foundations.} Our controller builds on classical bandit
tools: Thompson sampling \citep{thompson1933likelihood}, UCB \citep{auer2002finite}, and
confidence-based action elimination \citep{evendar2006action}, the last a direct precedent for
our retirement rule. We formulate the budgeted setting as bandits-with-knapsacks
\citep{badanidiyuru2018bandits}, adapting its linear-reward, resource-constrained framework to
a concave, diversity-gated value. The joint uncertainty-and-diversity objective also echoes
batch active learning \citep{ash2020deep}, though we act in a generative, budget-constrained
loop rather than selecting existing points to label.

\section{Method}
\label{sec:method}

\subsection{Problem setup and notation}
\label{sec:setup}

In this section, we define the general notation and problem setup of a bandit-with-knapsack algorithm\citep{badanidiyuru2018bandits}. A pool of $M$ problems or queries $\mathcal{Q}=\{q_1,\dots,q_M\}$ is drawn from a dataset equipped with an
automatic verifier (competition mathematics with integer answers, or programming problems
with unit tests). Given an input query, a teacher policy $\pi$ samples a trajectory $t$, extracts its final answer $\hat a$, verifies it's correctness as $y\in\{0,1\}$, and returns the prompt- and completion-token counts $c_{\mathrm{p}}$ and $c_{\mathrm{c}}$.
\paragraph{Cost and budget.} We measure the cost of generation, our budget, purely by the completion-token counts $c_{\mathrm{c}}$. Prompt tokens are fixed per query and identical across the policies we compare, so they contribute only a constant offset that cancels under a matched budget.
\paragraph{Arms and notation.} Each problem or query within the dataset denotes a bandit arm carrying a Beta posterior over the teacher's per-sample success probability $p_q$. Absent prior knowledge, a Beta-Bernoulli bandit starts from the uniform prior $\mathrm{Beta}(\alpha_0=1,\beta_0=1)$ and updates each arm's posterior as evidence arrives. After $n_q$ pulls of which $n_q^{+}$ are correct, the arm's posterior is $\mathrm{Beta}(\alpha_q,\beta_q)$, where $\alpha_q=\alpha_0+n_q^{+}$ counts the successful trials (verified-correct trajectories) and $\beta_q=\beta_0+(n_q-n_q^{+})$ counts the failures (verified-incorrect trajectories).
\paragraph{Objective.} We cast synthetic data collection as a bandits-with-knapsacks problem: maximize the total value collected across the pool subject to the fixed completion-token budget   \[
  \max\ \sum_{q\in\mathcal{Q}} V(q) \qquad\text{s.t.}\qquad \sum_{\text{pulls}} c_{\mathrm{c}} \le B,
  \]

\subsection{Overview}
\label{sec:overview}

\begin{figure}[t]
\centering
\resizebox{\textwidth}{!}{%
\begin{tikzpicture}[
  font=\footnotesize,
  box/.style={rounded corners=3pt, draw, line width=0.9pt, fill=white,
              blur shadow={shadow xshift=0.7pt, shadow yshift=-1.0pt, shadow opacity=35}},
  hdr/.style={anchor=north, minimum height=0.62cm, text=white, font=\small\bfseries,
              rounded corners=3pt},
  card/.style={rounded corners=2pt, draw=black!18, fill=white, line width=0.5pt},
  arr/.style={-{Stealth[length=3.2mm]}, line width=1.1pt, gray!60},
  flow/.style={-{Stealth[length=2.1mm]}, line width=0.8pt, black!45},
]
  \node[cylinder, shape border rotate=90, aspect=0.30, draw=gray!55, line width=0.9pt,
        fill=gray!10, minimum height=2.4cm, minimum width=2.1cm, align=center]
        (pool) at (0.7,5.5)
        {Query\\ pool $\mathcal{Q}$\\[2pt]
         {\normalsize\textcolor{easyc}{$\bullet$}\textcolor{medc}{$\bullet$}\textcolor{hardc}{$\bullet$}}\\[-2pt]
         {\scriptsize unknown}\\[-3pt]{\scriptsize difficulty}};

  \node[box, draw=calibc, minimum width=4.4cm, minimum height=8.4cm] (calib) at (4.9,5.5) {};
  \node[hdr, fill=calibc, minimum width=4.4cm] at (calib.north) {1\quad Calibration};

  \node[card, minimum width=3.9cm, minimum height=1.95cm] (cA) at (4.9,7.75) {};
  \node[font=\scriptsize\bfseries, anchor=north] at ([yshift=-2pt]cA.north)
        {difficulty prior $\mathrm{Beta}(\alpha_0,\beta_0)$};
  \begin{scope}[shift={($(cA.south west)+(0.5,0.28)$)}]
    \fill[calibc!18] plot[smooth,tension=0.6] coordinates
      {(0,0.03)(0.09,1.02)(0.28,0.66)(0.55,0.44)(0.95,0.27)(1.4,0.16)(1.95,0.08)(2.6,0.04)}
      -- (2.6,0) -- (0,0) -- cycle;
    \draw[calibc, very thick] plot[smooth,tension=0.6] coordinates
      {(0,0.03)(0.09,1.02)(0.28,0.66)(0.55,0.44)(0.95,0.27)(1.4,0.16)(1.95,0.08)(2.6,0.04)};
    \draw[gray!55] (0,0) -- (2.7,0);
    \node[font=\tiny, anchor=north] at (0,-0.01){0};
    \node[font=\tiny, anchor=north] at (2.6,-0.01){1};
    \node[font=\tiny, anchor=north, text=black!60] at (1.5,-0.02){$p$ (success rate)};
  \end{scope}

  \node[card, minimum width=3.9cm, minimum height=1.95cm] (cB) at (4.9,5.4) {};
  \node[font=\scriptsize\bfseries, anchor=north] at ([yshift=-2pt]cB.north)
        {generation\_len  $P_{95}\!\times\!$margin};
  \begin{scope}[shift={($(cB.south west)+(0.5,0.28)$)}]
    \foreach \x/\h in {0/0.18,0.32/0.42,0.64/0.82,0.96/1.02,1.28/0.7,1.6/0.4,1.92/0.2}
      \fill[gray!45] (\x,0) rectangle +(0.26,\h);
    \draw[gray!55] (0,0) -- (2.7,0);
    \draw[hardc, thick, dashed] (2.35,0) -- (2.35,1.15);
    \node[font=\tiny, text=hardc, anchor=south] at (2.42,0.95){cap};
    \node[font=\tiny, anchor=north, text=black!60] at (1.35,-0.02){completion length};
  \end{scope}

  \node[card, minimum width=3.9cm, minimum height=1.95cm] (cC) at (4.9,3.05) {};
  \node[font=\scriptsize\bfseries, anchor=north] at ([yshift=-2pt]cC.north)
        {diversity threshold $\tau$};
  \begin{scope}[shift={($(cC.south west)+(0.5,0.28)$)}]
    \fill[calibc!14] plot[smooth,tension=0.7] coordinates
      {(0,0.05)(0.35,0.75)(0.7,0.55)(1.1,0.18)(1.5,0.12)(1.9,0.5)(2.25,0.86)(2.6,0.4)}
      -- (2.6,0) -- (0,0) -- cycle;
    \draw[calibc, very thick] plot[smooth,tension=0.7] coordinates
      {(0,0.05)(0.35,0.75)(0.7,0.55)(1.1,0.18)(1.5,0.12)(1.9,0.5)(2.25,0.86)(2.6,0.4)};
    \draw[gray!55] (0,0) -- (2.7,0);
    \draw[medc, thick, dashed] (1.3,0) -- (1.3,0.95);
    \node[font=\tiny, text=medc, anchor=south] at (1.3,0.95){$\tau$};
    \node[font=\tiny, anchor=north, text=black!60] at (0.6,-0.02){novel};
    \node[font=\tiny, anchor=north, text=black!60] at (2.25,-0.02){same};
  \end{scope}

  \begin{scope}[xshift=1.9cm]
  \node[box, draw=sampc, minimum width=7.8cm, minimum height=8.4cm] (samp) at (11.6,5.5) {};
  \node[hdr, fill=sampc, minimum width=7.8cm] at (samp.north)
        {2\quad Sampling controller\; \mdseries\small(bandits-with-knapsacks)};

  \node[card, minimum width=2.9cm, minimum height=2.0cm] (arms) at (9.35,7.6) {};
  \node[font=\scriptsize\bfseries, anchor=north] at ([yshift=-2pt]arms.north)
        {arm posteriors $+\,\tilde p_q$};
  \begin{scope}[shift={($(arms.south west)+(0.35,0.30)$)}]
    \fill[hardc!20] plot[smooth,tension=0.6] coordinates{(0.05,0.05)(0.22,0.62)(0.4,0.98)(0.6,0.6)(0.85,0.18)(1.05,0.05)}--cycle;
    \draw[hardc,thick] plot[smooth,tension=0.6] coordinates{(0.05,0.05)(0.22,0.62)(0.4,0.98)(0.6,0.6)(0.85,0.18)(1.05,0.05)};
    \fill[medc!22] plot[smooth,tension=0.6] coordinates{(0.75,0.05)(0.98,0.4)(1.25,0.72)(1.5,0.4)(1.72,0.1)}--cycle;
    \draw[medc,thick] plot[smooth,tension=0.6] coordinates{(0.75,0.05)(0.98,0.4)(1.25,0.72)(1.5,0.4)(1.72,0.1)};
    \fill[easyc!22] plot[smooth,tension=0.6] coordinates{(1.55,0.04)(1.8,0.34)(2.1,0.66)(2.35,0.34)(2.55,0.08)}--cycle;
    \draw[easyc,thick] plot[smooth,tension=0.6] coordinates{(1.55,0.04)(1.8,0.34)(2.1,0.66)(2.35,0.34)(2.55,0.08)};
    \draw[gray!55] (0,0) -- (2.65,0);
    \draw[hardc!70, densely dotted] (0.34,0)--(0.34,0.55);
    \draw[medc!70, densely dotted] (1.32,0)--(1.32,0.62);
    \draw[easyc!70, densely dotted] (2.02,0)--(2.02,0.56);
    \node[font=\tiny, anchor=north, text=black!60] at (1.3,-0.02){$p$};
  \end{scope}

  \node[card, draw=sampc, line width=0.9pt, fill=sampc!10, minimum width=3.55cm,
        minimum height=2.3cm] (score) at (13.4,7.55) {};
  \node[font=\scriptsize\bfseries, text=sampc, anchor=north] at ([yshift=-3pt]score.north)
        {diversity-aware selection};
  \node[anchor=north, font=\footnotesize, align=center] at ([yshift=-17pt]score.north)
        {$\mathrm{score}(q)=$\\[4pt]
         $\underbrace{\tilde p_q}_{\text{Thomp.}}\,
          \underbrace{q_{\mathrm{nov}}}_{\text{nov.}}\,
          \underbrace{\tfrac{1}{1{+}v}}_{\text{conc.}}\,
          \underbrace{\tfrac{1}{c_q}}_{\text{cost}}$};

  \node[card, minimum width=2.9cm, minimum height=2.0cm] (gate) at (9.35,5.0) {};
  \node[font=\scriptsize\bfseries, anchor=north] at ([yshift=-2pt]gate.north)
        {diversity gate $<\tau$};
  \begin{scope}[shift={($(gate.south west)+(0.28,0.30)$)}]
    \foreach \y in {0.15,0.55,0.95} \fill[gray!35] (0,\y) rectangle +(0.16,0.16);
    \draw[sampc!60, fill=sampc!8] (0.5,0.15) -- (0.95,0.45) -- (0.95,0.85) -- (0.5,1.15) -- cycle;
    \fill[easyc] (1.35,0.9) rectangle +(0.16,0.16);
    \fill[medc]  (1.57,0.9) rectangle +(0.16,0.16);
    \node[font=\tiny, anchor=west, text=black!70] at (1.8,0.98){$v(q){=}2$};
    \draw[gray!55] (1.35,0.25) rectangle +(0.16,0.16);
    \draw[gray!55] (1.57,0.25) rectangle +(0.16,0.16);
    \node[font=\tiny, text=hardc] at (1.43,0.33){\ttfamily x};
    \node[font=\tiny, text=hardc] at (1.65,0.33){\ttfamily x};
    \node[font=\tiny, anchor=west, text=black!55] at (1.8,0.33){dup.};
  \end{scope}

  \node[card, minimum width=3.55cm, minimum height=2.0cm] (alloc) at (13.4,5.0) {};
  \node[font=\scriptsize\bfseries, anchor=north] at ([yshift=-2pt]alloc.north)
        {per-token efficiency};
  \begin{scope}[shift={($(alloc.south west)+(0.55,0.32)$)}]
    \draw[gray!55] (0,0) -- (2.5,0);      
    \draw[gray!55] (0,0) -- (0,1.28);     
    \draw[sampc, very thick] plot[smooth,tension=0.7] coordinates
      {(0.05,1.20)(0.55,0.97)(1.15,0.79)(1.75,0.66)(2.35,0.57)};
    \draw[gray!65, very thick] plot[smooth,tension=0.7] coordinates
      {(0.05,0.90)(0.55,0.69)(1.15,0.55)(1.75,0.46)(2.35,0.40)};
    \node[font=\tiny, anchor=north, text=black!60] at (1.25,-0.03){token budget};
    \node[font=\tiny, rotate=90, anchor=south, text=black!60] at (-0.12,0.64){dv/1k};
    \node[font=\tiny, text=sampc] at (1.55,1.06){THRIFTY};
    \node[font=\tiny, text=black!55] at (1.75,0.52){uniform};
  \end{scope}

  \node[card, draw=hardc!35, minimum width=7.1cm, minimum height=2.15cm] (ret) at (11.55,2.75) {};
  \node[font=\scriptsize\bfseries, text=hardc!85, anchor=north] at ([yshift=-2pt]ret.north)
        {opportunity-cost retirement};
  \begin{scope}[shift={($(ret.west)+(0.5,-0.35)$)}]
    \fill[hardc!14] plot[smooth,tension=0.6] coordinates{(0,0.04)(0.12,0.75)(0.32,0.5)(0.6,0.28)(1.0,0.12)(1.5,0.05)}--(1.5,0)--(0,0)--cycle;
    \draw[hardc,very thick] plot[smooth,tension=0.6] coordinates{(0,0.04)(0.12,0.75)(0.32,0.5)(0.6,0.28)(1.0,0.12)(1.5,0.05)};
    \draw[hardc!40,thick,densely dashed] plot[smooth,tension=0.6] coordinates{(0,0.05)(0.35,0.45)(0.75,0.5)(1.1,0.35)(1.5,0.15)};
    \draw[gray!55] (0,0)--(1.7,0);
    \draw[black!55, dashed] (0.55,0)--(0.55,0.7);
    \node[font=\tiny, anchor=south, text=black!70] at (0.55,0.68){$p_{\min}$};
    \draw[flow, black!45] (0,-0.12) -- node[font=\tiny,below,black!55]{failures}(1.5,-0.12);
  \end{scope}
  \begin{scope}[shift={($(ret.center)+(0.75,-0.35)$)}]
    \draw[hardc!80, thick, dashed] (0,0.45) -- (2.9,0.45);
    \node[font=\tiny, text=hardc, anchor=east] at (2.9,0.58){waterline $\lambda$};
    \fill[easyc] (0.3,0.85) circle (0.07); \fill[medc] (0.9,0.72) circle (0.07);
    \fill[easyc] (1.5,0.9) circle (0.07);  \fill[medc] (2.05,0.66) circle (0.07);
    \fill[hardc] (1.2,0.16) circle (0.07);
    \node[font=\tiny, text=hardc] at (1.42,0.16){\ttfamily x};
    \node[font=\tiny, anchor=north, text=black!60] at (1.2,-0.02){retired ($<\lambda$)};
  \end{scope}

  \draw[flow] (arms.east) -- (score.west);
  \draw[flow] (score.south) .. controls +(0,-0.5) and +(0,0.5) .. (alloc.north);
  \draw[flow] (gate.east) -- (alloc.west);

  \node[box, draw=outc, minimum width=3.6cm, minimum height=8.4cm] (out) at (17.9,5.5) {};
  \node[hdr, fill=outc, minimum width=3.6cm] at (out.north) {3\quad Collected data};

  \node[card, minimum width=3.1cm, minimum height=1.95cm] (traces) at (17.9,7.6) {};
  \node[font=\scriptsize\bfseries, text=outc, anchor=north] at ([yshift=-2pt]traces.north)
        {distinct correct traces};
  \begin{scope}[shift={($(traces.center)+(-0.55,-0.55)$)}]
    \fill[outc!12,draw=outc!60] (0.2,0.15) rectangle +(0.8,1.0);
    \fill[outc!12,draw=outc!60] (0.1,0.05) rectangle +(0.8,1.0);
    \fill[white,draw=outc!70,line width=0.7pt] (0,-0.05) rectangle +(0.8,1.0);
    \foreach \y in {0.15,0.35,0.55,0.75} \draw[outc!55] (0.12,\y)--(0.68,\y);
    \node[font=\tiny, anchor=west, text=black!60] at (1.05,0.45){for SFT};
  \end{scope}

  \node[card, minimum width=3.1cm, minimum height=1.55cm] (dm) at (17.9,5.25) {};
  \node[font=\scriptsize\bfseries, text=outc, anchor=north] at ([yshift=-2pt]dm.north)
        {free difficulty model};
  \node[font=\scriptsize, anchor=north, align=center] at ([yshift=-15pt]dm.north)
        {per-query posterior\\ $\hat p_q=\alpha_q/(\alpha_q{+}\beta_q)$};

  \node[card, draw=black!35, dashed, fill=black!3, minimum width=3.1cm, minimum height=2.4cm] (stud) at (17.9,2.75) {};
  \node[font=\scriptsize\bfseries, text=black!45, anchor=north] at ([yshift=-2pt]stud.north)
        {student SFT};
  \node[font=\tiny\itshape, text=black!40, anchor=north] at ([yshift=-11pt]stud.north)
        {(downstream, out of scope)};
  \begin{scope}[shift={($(stud.center)+(-1.05,-0.62)$)}]
    \fill[black!45] (0,0) rectangle +(0.4,0.42);
    \fill[black!32] (0.5,0) rectangle +(0.4,0.42);
    \fill[black!45] (1.0,0) rectangle +(0.4,0.42);
    \fill[black!55] (1.5,0) rectangle +(0.4,0.42);
    \draw[flow, black!40] (0,-0.18) -- (1.9,-0.18);
    \node[font=\tiny, anchor=north, text=black!45] at (0.95,-0.2){easy $\rightarrow$ hard curriculum};
  \end{scope}
  \draw[flow, black!35, dashed] (dm.south) -- (stud.north);
  \end{scope}

  \draw[arr] (pool.east) -- node[font=\scriptsize, above, black!60]{probe}(calib.west);
  \draw[arr] (calib.east) -- node[font=\scriptsize, align=center, above, black!60]
        {$(\alpha_0,\beta_0),\tau,L_{\max}$}
        node[font=\scriptsize, align=center, below, black!60]{warm-start}(samp.west);
  \draw[arr] (samp.east) -- node[font=\scriptsize, above, black!60]{collect}(out.west);
\end{tikzpicture}%
}
  \caption{The THRIFTY pipeline. \textbf{Stage 1 (Calibration)} probes a small subset (5--10\%) of the dataset to derive sensible
  presets (generation length, query difficulty, diversity threshold etc.) from data rather
  than hand-tuning them. \textbf{Stage 2 (Sampling)} each round samples the problems of highest
  expected value per token, trading off solvability, solution novelty, diminishing returns, and cost. The net policy
  undersamples easy problems, oversamples the hard-but-solvable, and retires the hopeless. The run
  yields the distinct correct traces together with a per-query difficulty model at no extra cost.
  The greyed \emph{student SFT} block illustrates a downstream use of this data (curriculum training
  a student) that is outside the scope of this paper.}
  \label{fig:pipeline}
\end{figure}
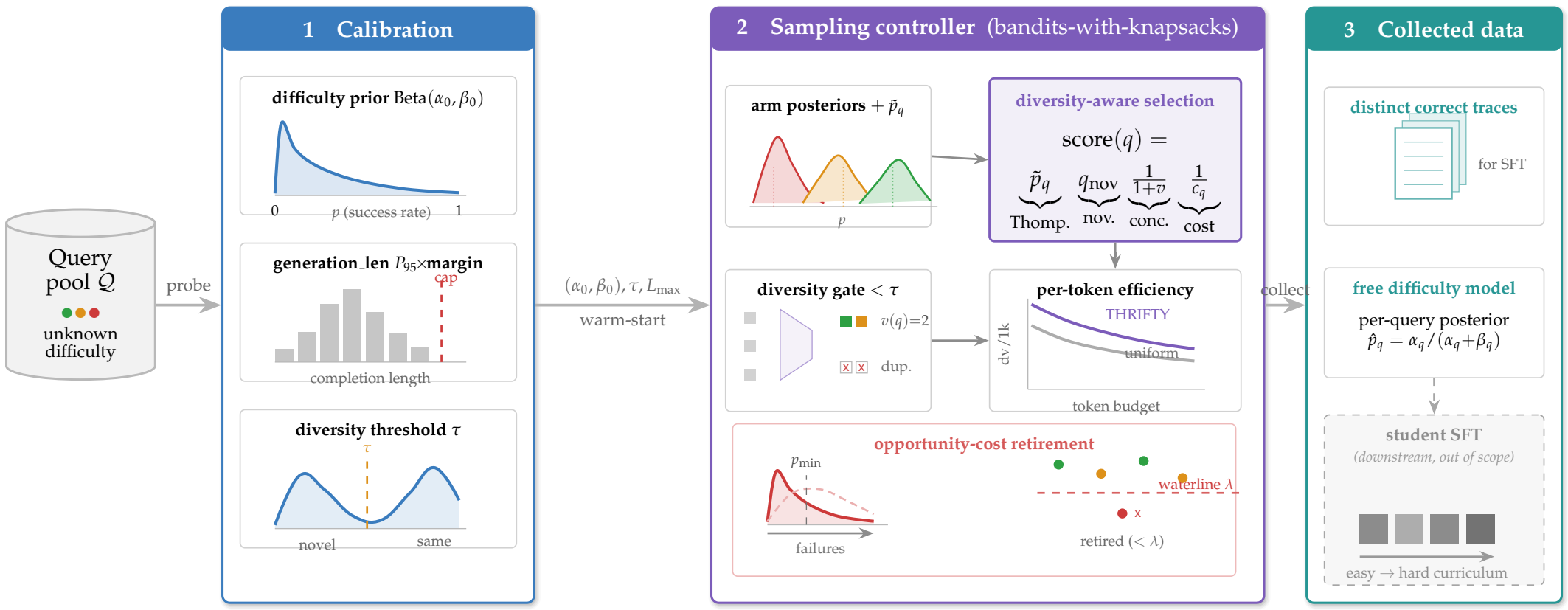

We now introduce THRIFTY, which builds on the standard bandits-with-knapsacks formulation above and improves it in the following ways.

Our data generation controller runs in two stages(Figure~\ref{fig:pipeline}): a \emph{\textbf{calibration stage}}, and a \emph{\textbf{sampling stage}}, both of which consume budget from a single pool of global token budget $B$. The goal of the calibration stage is to empirically obtain sensible presets such as the max generation length, dataset level difficulty priors which encode the generating model's overall capability on the task, and diversity thresholds over the sampled trajectories. In a traditional data generation pipeline, these presets would be either set by trial and error or decided based on some domain knowledge. With our approach, we spend some budget in this stage, but obtain hard empirical presets in a closed form fashion purely from the data, with the goal that the efficiency obtained for the next sampling stage, would offset the small cost incurred in obtaining these parameters.

The \emph{sampling} stage then runs a policy-agnostic controller loop to the remaining budget,
driven by two coupled decisions: an allocation policy that spends the next token on the arm of
highest marginal value per token, and a retirement rule that removes an arm once the evidence
says further budget is better spent elsewhere. The run leaves behind both a dataset of
distinct correct trajectories and, at no additional cost, an extremely valuable per-query posterior which serves as a difficulty model of the query pool.

\subsection{Stage 1: Calibration}
\label{sec:calib}

Calibration spends a small fixed fraction of $B$ to replace hand-guessed knobs with data-derived ones. The calibration stage ingests a subset(typically 5-10 \%, a design choice) from the overall random pool of queries in the datasets and produces the following outputs:

\paragraph{(a) Dataset Difficulty prior.} We sample a handful of pulls on a calibration subset, form each probed problem’s empirical success rate, and moment-match a dataset level $\mathrm{Beta}(\alpha_0,\beta_0)$ prior. This gives us  a more informative initialization for the Beta-Bernoulli bandit rather than just initializing it with a uniform prior $\mathrm{Beta}(\alpha_0=1,\beta_0=1)$. This empirically obtained prior although obtained over a random subset of the whole dataset, serves as a good proxy for the overall capability of the teacher/generating model as well the task difficulty.

\paragraph{(b) Max Generation-length} The teacher's maximum generation length is set to a high percentile(p95) of the observed completion-length distribution times a safety margin to accomodate any long tails in the query length.
$L_{\max} = \lceil m \cdot P_{95} \rceil$, for our experiment we set this safety margin m as 20\%(m=1.2) over the 95 percentile comepltion lengths observed. Since some ceiling must be set even during calibration, we use a generous one (4096 tokens) and, whenever a rollout
hits it, raise it and re-probe, so the fitted cap is not biased by truncation. In practice, no rollout hit this ceiling on any of our datasets in our experiments, so re-probing was never needed.

\paragraph{(c) Diversity threshold $\tau$.} The diversity metric is itself a design choice that can be swapped per task; for our experiments we use self-BLEU \citep{zhu2018texygen} to quantify how similar two reasoning trajectories are. The single knob of
the diversity gate (Section~\ref{sec:alloc}) is then the self-BLEU boundary $\tau$ above which two correct traces count
as the same solution method. We pool, over the calibration set, the pairwise self-BLEU between correct traces of the \emph{same} problem, and place $\tau$ in the low-density valley between the two modes of that distribution, locating the valley by Otsu's method \citep{otsu1979}, a standard histogram-thresholding criterion that chooses the split maximizing the between-class variance of the two modes. If the diversity distribution doesn't have any separating modes, it falls back to a default value of $\tau=0.5$.

\paragraph{(d) Warm-start as continuation.} Correct trajectories produced during calibration are
not discarded. They carry into the sampling stage as real progress: they count toward each
problem's value and appear in the final dataset, so no probe token is wasted. An arm that has
calibration data continues from its updated posterior; only arms with no calibration data start
from the global dataset prior $\mathrm{Beta}(\alpha_0,\beta_0)$ defined in a). Because we charge the probe cost against $B$, the
budget we report already includes calibration, and our experiments show that cost is amortized by
the efficiency the fitted settings buy.
\subsection{Stage 2, part I: Diversity-gated knapsack allocation}
\label{sec:alloc}

Instead of just maximizing the count of correct trajectories over the dataset, we formulate the bandit's value function to reward correct \emph{and distinct}
solutions under diminishing returns, spending each token where its marginal value per token is highest, the bandits-with-knapsacks view of a fixed resource spent by marginal utility \citep{badanidiyuru2018bandits}.

\paragraph{Diversity as a gate.} We use self-BLEU as a diversity gate: a newly sampled correct trajectory is accepted only if its self-BLEU(4-gram) against the
query's already-accepted traces falls below the calibrated threshold $\tau$, keeping only distinct solutions per query. As stated earlier, the metric itself is a design choice and can be swapped in our formulation as long as a corresponding gating mechanism for novelty is identified.

\paragraph{A concave value and its per-token score.} We define the value function in the bandits optimization objective as a concave function, like below The per-problem value is concave in the
distinct count,
\[
\boxed{\,V(q)=\ln\!\big(1+v(q)\big)\,},
\]
with $V(0)=0$ and diminishing marginal returns. Because $V$ is concave, the budget-constrained
optimum of $\sum_q V(q)$ is a shadow-price policy: spend the next token where marginal value per
token is highest and stop when the budget is gone; we never compute the shadow price, since
running to budget realizes it. A pull banks a new distinct solution only if it is both correct and
novel, raising value by $\Delta V(v)=\ln(2+v)-\ln(1+v)\approx 1/(1+v)$; dividing the expected gain
by the pull's token cost gives the selection score (derivation in Appendix~\ref{app:score}):
\[
\boxed{\;\mathrm{score}(q) \;=\; \tilde p_q \cdot q_{\mathrm{novel}}(q)\cdot \frac{1}{1+v(q)}\cdot \frac{1}{c_q}\;}
\]
Each factor is independently interpretable: the Thompson draw
$\tilde p_q\sim\mathrm{Beta}(\alpha_q,\beta_q)$ of solvability \citep{thompson1933likelihood},
which preserves the exploration a point estimate would lose; the novelty ratio
$q_{\mathrm{novel}}\!=\!v/n_q^{+}$ (one before the first correct trace); the concave discount
$1/(1+v)$; and the cost term $1/c_q$ with $c_q$ the running mean completion tokens per pull.
Relative to a plain Thompson score $\tilde p_q$, the objective adds only two multiplicative
factors and a cost normalizer.

\subsection{Stage 2, part II: opportunity-cost retirement}
\label{sec:retire}
Even a well-tuned allocation policy keeps leaking budget on problems the teacher simply cannot
solve. Thompson sampling only gives up on an arm gradually: until its posterior collapses toward
zero, an occasional optimistic draw still spends tokens on it, and those tokens buy nothing. A
retirement criterion cuts a hopeless problem off explicitly, well before the policy would starve
it on its own, and frees that budget for problems still worth attempting. We take inspiration from
the opportunity-cost retirement principle of bandits-with-knapsacks \citep{badanidiyuru2018bandits}
and confidence-based action elimination \citep{evendar2006action}: a problem is dropped only once
its budget is better spent elsewhere.
Each round we ask how many more pulls the leftover budget can afford,
$K_{\mathrm{afford}}=\text{remaining}/\bar c$ at an average cost $\bar c$ per pull, and read off
the value-per-token of the $K_{\mathrm{afford}}$-th most promising pull still available; this is
the \emph{waterline} $\lambda$, and any pull worth less than $\lambda$ is one we can no longer
afford. A problem is retired once it has returned no correct trace after a few pulls and even its
\emph{best-case} value per token falls below $\lambda$; using the best case rather than the
average keeps barely-tried problems safe and retires only those beaten down by repeated failure.
The waterline
tracks the budget automatically: when budget is ample it clears every active problem and nothing
is retired; as budget runs low $\lambda$ rises and the clearly-dead problems fall below it.

Algorithm~\ref{alg:thrifty} ties these pieces together into a single run: starting from the
calibrated warm-start, each round scores the active arms, samples a batch of the highest-value
problems, updates their posteriors, admits only the correct traces that clear the diversity gate,
and retires arms that fall below the waterline, repeating until the budget is spent.

\begin{algorithm}[h]
\caption{THRIFTY sampling controller (one run)}
\label{alg:thrifty}
\begin{algorithmic}[1]
\Require budget $B$, batch size $b$, prior $(\alpha_0,\beta_0)$, diversity threshold $\tau$, retirement floor $n_{\min}$, best-case level $\rho$
\State initialize each query arm q from its calibration warm-start or from the global dataset prior; $\mathrm{spent}\gets$ calibration tokens
\While{$\mathrm{spent} < B$ \textbf{and} at least one arm is active}
  \State recompute waterline $\lambda$ from the remaining budget $B-\mathrm{spent}$; retire any active arm with $n_q^{+}=0$, $n_q\ge n_{\min}$, whose best-case value density (level $\rho$) is $<\lambda$
  \State $\mathcal{A}\gets\{q : \lnot\,\mathrm{retired}(q)\}$; \quad compute $\mathrm{score}(q)$ for all $q\in\mathcal{A}$
  \State select the $b$ highest-scoring distinct arms $\mathcal{B}\subseteq\mathcal{A}$
  \State issue $\texttt{generate}(q)$ concurrently for all $q\in\mathcal{B}$
  \For{each returned $(t,\hat a,y,c_{\mathrm{c}})$ on arm $q$}
     \State $\mathrm{spent}\mathrel{+}=c_{\mathrm{c}}$;\quad update running cost $c_q$;\quad update posterior $(\alpha_q,\beta_q)$ with $y$
     \If{$y=1$ \textbf{and} $\mathrm{selfBLEU}(t;\ \text{accepted}_q)<\tau$}
        \State accept $t$;\quad $v(q)\mathrel{+}=1$
     \EndIf
     \State update $q_{\mathrm{novel}}(q)$
  \EndFor
\EndWhile
\State \Return accepted distinct trajectories and per-arm posteriors $\{\mathrm{Beta}(\alpha_q,\beta_q)\}$
\end{algorithmic}
\end{algorithm}

\section{Experiments}
\label{sec:exp}

\subsection{Experimental setup}
\label{sec:exp-setup}

\textbf{Datasets.} Given limited compute capacity, we evaluate our method on subsets of popular math (AIME 300,
  MATH-500) and code (LiveCodeBench) benchmarks, chosen to span both difficulty and diversity
  headroom. For \emph{AIME}, we reuse the exact publicly available pool of $300$ randomly sampled
  problems from prior work \citep{fastmcts}. For \emph{MATH-500}, we use the full $500$-problem test
  split of \citet{lightman2024lets}. For \emph{LiveCodeBench} (LCB) \citep{jain2024livecodebench},
  we use the complete standard-I/O subset of $183$ problems. We provide all these subsets in our code release for reproducibility.

\textbf{Teacher model and Budget.} We use the Qwen2.5-32B-Instruct as our Teacher or policy model for the data generation. We define three budget regimes corresponding to a uniform fixed number of pulls K per query across the entire dataset, such that $K\in\{2,4,8\}$ corresponds to the (scarce/medium/ample) budget regime. The token budget for the experiment is thus calculated $B=K\cdot M\cdot\bar c$ with $M$ representing the dataset  size and $\bar c$ the mean completion length.  We want to stress that choosing a fixed token budget(say 500K, 1M, 2M tokens) would not be comparable across our datasets, as they are of varying sizes. For eg, in the case of LiveCodeBench consisting of 183 queries, 500K budget itself amounts to an ample regime($\approx$ 8.4 pulls/query), as the avg completion length is 387.

For calibration stage, we utilize a random 10\% subset of the total dataset, with at-most $K$ pulls ($K\in\{2,4,8\}$ corresponding to scarce/medium/ample regimes). While this doesn't lead to the calibration stage translate to a 10\% of the total budget since the calibration stage doesn't cap generation length, we find this to be a good enough approximation to allocate the calibration budget. In practice it consumes only $9$--$12\%$ of the total budget across datasets and budget regimes (Table~\ref{tab:calib}). This probe cost is included in the total budget, so for our method the available generation budget is the total minus the calibration budget ($B_{\mathrm{gen}} = B - B_{\mathrm{calib}}$).

\begin{table}[t]
\centering
\footnotesize
\setlength{\tabcolsep}{7pt}
\caption{\textbf{Calibration cost.} Tokens spent by the calibration stage (Stage~1, incurred only
by THRIFTY; the baselines run without calibration) and its share of the total budget $B$, per
dataset and budget regime. Calibration probes a random
$10\%$ subset of the pool at $K$ pulls per query and consistently consumes only $9$--$12\%$ of
$B$; this probe cost is consumed from the same total budget.}
\label{tab:calib}
\begin{tabular}{l r r r}
\toprule
Budget regime & Budget $B$ & Calib.\ tokens & \% of $B$\\
\midrule
\multicolumn{4}{l}{\emph{AIME} ($M{=}300$; probe $=30$ queries, $10\%$)}\\
\quad scarce ($K{=}2$)  & $540$k  & $52.5$k  & $9.7\%$\\
\quad medium ($K{=}4$)  & $1.09$M & $109.3$k & $10.0\%$\\
\quad ample ($K{=}8$)   & $2.18$M & $218.7$k & $10.0\%$\\
\addlinespace[2pt]
\multicolumn{4}{l}{\emph{MATH-500} ($M{=}500$; probe $=50$ queries, $10\%$)}\\
\quad scarce ($K{=}2$)  & $570$k  & $61.3$k  & $10.8\%$\\
\quad medium ($K{=}4$)  & $1.15$M & $123.8$k & $10.8\%$\\
\quad ample ($K{=}8$)   & $2.30$M & $247.6$k & $10.8\%$\\
\addlinespace[2pt]
\multicolumn{4}{l}{\emph{LiveCodeBench} ($M{=}183$; probe $=18$ queries, $10\%$)}\\
\quad scarce ($K{=}2$)  & $130$k  & $13.8$k  & $10.6\%$\\
\quad medium ($K{=}4$)  & $260$k  & $27.5$k  & $10.6\%$\\
\quad ample ($K{=}8$)   & $520$k  & $60.5$k             & $11.6\%$\\
\bottomrule
\end{tabular}
\end{table}

\textbf{Metrics.} To paint a holistic, balanced picture of evaluation across dataset breadth,
depth, and diversity, we define the following three metrics, all averaged over seeds:
\begin{itemize}\setlength{\itemsep}{2pt}\setlength{\parskip}{0pt}
  \item \textbf{dv/1k (diverse value per 1k tokens)} --- our primary, efficiency metric. Let
  $v(q)$ be the number of distinct correct trajectories accepted for problem $q$ (those that pass
  the self-BLEU novelty gate), and let the collected value be $V=\sum_{q}\ln\!\big(1+v(q)\big)$,
  the objective itself. We divide this by the total budget tokens the run consumes, $T=\sum_q c(q)$,
  and rescale to a per-thousand-token rate,
  \[
    \mathrm{dv/1k} \;=\; \frac{1000\,V}{T} \;=\; \frac{1000\sum_{q}\ln(1+v(q))}{\sum_q c(q)}.
  \]
  It measures the diversity-gated value bought per token and is comparable across datasets and
  budget regimes; it is the fair way to score allocators that spend unequally (the cost thesis of
  the paper).
  \item \textbf{coverage@1} --- the number of problems solved at least once (at least one verified
  correct trace). It measures breadth: whether the allocator finds the solvable problems, and acts
  as a guardrail that efficiency is not bought by sacrificing coverage.
  \item \textbf{total distinct trajectories} --- the raw count $\sum_q v(q)$ of accepted distinct
  correct traces, i.e.\ the volume of usable SFT data the run collects.
\end{itemize}

\subsection{Baselines}
\label{sec:baselines}

All allocators share the identical policy-agnostic controller loop and the
  same budget $B$. Crucially, the diversity gate and value $\ln(1+v)$ are applied when scoring
  \emph{every} allocator, so they differ only in how they \emph{allocate}, not how they are
  \emph{measured}. We compare two families.

  \smallskip\noindent\textit{Static allocation} (budget- and calibration-agnostic; flat
  $\mathrm{Beta}(1,1)$ prior, fixed generation length, no warm-start):
  \begin{itemize}\setlength{\itemsep}{1pt}\setlength{\parskip}{0pt}
    \item \emph{Random.} Each round it picks a query uniformly at random (with replacement) and
    samples it once, until $B$ is spent. Selection is stochastic, so the per-problem sample counts
    are balanced only in expectation and fluctuate across problems; it is the naive lower bound.
    \item \emph{Uniform.} Round-robin over the full pool: it cycles through every problem in turn,
    so all problems receive an (essentially) equal number of samples by construction---the standard
    rejection-sampling recipe. The two differ only in how they spread the budget: \emph{Uniform}
    is deterministic and exactly even, while \emph{Random} is stochastic and even only in
    expectation. Neither is difficulty-aware, so both keep sampling easy and unsolvable problems as
    heavily as hard-but-solvable ones.
  \end{itemize}

  \smallskip\noindent\textit{Dynamic allocation} (baselines). These are also bandit allocators,
  adapting their choice online from each arm's Beta posterior, and differ only in the fixed
  exploration rule they borrow from prior work. Each is started from a flat $\mathrm{Beta}(1,1)$,
  without calibration or warm-start, so their complete budget is devoted to generation and no
  calibration cost is incurred:
  \begin{itemize}\setlength{\itemsep}{2pt}\setlength{\parskip}{0pt}
    \item \emph{Difficulty-aware Bandit.} A curriculum / learning-progress baseline
    \citep{graves2017automated,bae2025online} that concentrates budget on problems of intermediate
    difficulty. It scores each query by the mid-difficulty band $w(q)=\hat p_q(1-\hat p_q)$, which
    peaks at $\hat p_q\!=\!0.5$, using the posterior-\emph{mean} success rate $\hat p_q$ rather than
    a Thompson draw, and selects by \emph{probability-matching}---query $q$ is sampled with
    probability $w(q)/\sum_{q'}w(q')$---rather than $\arg\max$, which would starve coverage by
    collapsing onto a few near-$0.5$ queries.
    \item \emph{Entropy Bandit.} The entropy exploration rule of \citet{zuo2025strategic}, taken
    directly from their paper. Each round it samples the query
    \[
      q_t \;=\; \arg\max_{q}\;\; \underbrace{H(\hat a_q)}_{\text{answer entropy}} \;+\;
      \lambda\,\underbrace{N_q^{-1/2}}_{\text{exploration bonus}},
    \]
    where $H(\hat a_q)$ is the Shannon entropy of the empirical distribution over the extracted
    answers seen so far for $q$, $N_q$ is its number of pulls, and $\lambda$ weights an exploration
    bonus that decays as the query is sampled. It thus favours queries whose sampled answers stay
    diverse (high entropy) rather than collapsing to a single value.
  \end{itemize}

  \smallskip\noindent Our method, \textbf{THRIFTY}, is the third bandit allocator. Unlike the two
  baselines above it is calibrated and warm-started, and it replaces their fixed exploration rules
  with a Thompson marginal-value score $\tilde p_q\,q_{\mathrm{novel}}\,\tfrac{1}{1+v}\,\tfrac{1}{c_q}$
  coupled to opportunity-cost retirement (Sections~\ref{sec:alloc}--\ref{sec:retire}).

\subsection{Results}

Table~\ref{tab:main} reports per-token efficiency (dv/1k) for all five allocators across the three
budget regimes, and Table~\ref{tab:allmetrics} breaks all three metrics down by regime for the
standard recipe, the closest prior bandit, and THRIFTY. THRIFTY attains the best dv/1k in eight of
the nine cells---tying Uniform on AIME-scarce, within seed noise---with its clearest edges on
MATH-500 ($\sim{+}11\%$) and LiveCodeBench ($\sim{+}15\%$), and the edge persists into the ample
regime rather than washing out. The gain is genuinely
per-token: THRIFTY matches the baselines' coverage@1 in every cell, so breadth is not sacrificed,
while the distinct-trajectory count trades off by dataset---higher on code, lower on the
answer-collapsed math pools---tracking where diverse solution paths actually exist.

Notably, Uniform is a stronger baseline than expected, beating both adaptive bandit baselines in 5 of 9 cells. This is because our pools are largely solvable, so evenly sampling every problem
already harvests most of the collectable value; the adaptive baselines instead misallocate—Difficulty-aware Bandit concentrates on mid-difficulty problems and starves the many easy ones that cheaply yield correct traces, while Entropy blindly targets answer-diversity that is particularly weak on the answer-collapsed math pools (thus helping only on code benchmark). THRIFTY retains the Uniform's coverage while adding value- and cost-aware allocation, so it stays ahead of both.

\begin{table}[t]
\centering
\footnotesize
\setlength{\tabcolsep}{4pt}
\caption{\textbf{Main results: per-token efficiency (dv/1k).} Diverse value per $1000$ budget
tokens, $\mathrm{dv/1k}=1000\sum_q\ln(1+v(q))/\sum_q c(q)$ (higher is better). Cells are
mean$_{\pm\text{std}}$ over $3$ seeds; budgets follow $B=K\cdot M\cdot\bar c$ completion
tokens ($K{=}2,4,8$ are the scarce/medium/ample regimes, i.e.\ $25/50/100\%$ of the ample budget).
Best per column in bold.}
\label{tab:main}
\resizebox{\textwidth}{!}{%
\begin{tabular}{l ccc ccc ccc}
\toprule
& \multicolumn{3}{c}{AIME ($M{=}300$)} & \multicolumn{3}{c}{MATH-500 ($M{=}500$)} & \multicolumn{3}{c}{LiveCodeBench ($M{=}183$)}\\
\cmidrule(lr){2-4}\cmidrule(lr){5-7}\cmidrule(lr){8-10}
Allocator & $K{=}2$ & $K{=}4$ & $K{=}8$ & $K{=}2$ & $K{=}4$ & $K{=}8$ & $K{=}2$ & $K{=}4$ & $K{=}8$\\
\midrule
\multicolumn{10}{l}{\emph{Static allocation (baselines)}}\\
\quad Random              & 0.160$_{\pm0.007}$ & 0.123$_{\pm0.003}$ & 0.095$_{\pm0.005}$ & 0.589$_{\pm0.010}$ & 0.387$_{\pm0.012}$ & 0.254$_{\pm0.013}$ & 0.463$_{\pm0.001}$ & 0.349$_{\pm0.005}$ & 0.148$_{\pm0.011}$ \\
\quad Uniform              & \textbf{0.179}$_{\pm0.004}$ & 0.126$_{\pm0.002}$ & 0.089$_{\pm0.003}$ & 0.676$_{\pm0.001}$ & 0.397$_{\pm0.009}$ & 0.233$_{\pm0.008}$ & 0.535$_{\pm0.029}$ & 0.348$_{\pm0.028}$ & 0.160$_{\pm0.021}$ \\
\addlinespace[2pt]
\multicolumn{10}{l}{\emph{Dynamic allocation (baselines)}}\\
\quad Difficulty-aware Bandit \citep{graves2017automated} & 0.170$_{\pm0.003}$ & 0.122$_{\pm0.004}$ & 0.088$_{\pm0.004}$ & 0.632$_{\pm0.007}$ & 0.403$_{\pm0.013}$ & 0.257$_{\pm0.012}$ & 0.527$_{\pm0.025}$ & 0.340$_{\pm0.015}$ & 0.150$_{\pm0.016}$ \\
\quad Entropy Bandit \citep{zuo2025strategic} & 0.172$_{\pm0.002}$ & 0.118$_{\pm0.002}$ & 0.081$_{\pm0.003}$ & 0.671$_{\pm0.007}$ & 0.397$_{\pm0.006}$ & 0.235$_{\pm0.010}$ & 0.547$_{\pm0.000}$ & 0.397$_{\pm0.000}$ & 0.151$_{\pm0.009}$ \\
\addlinespace[2pt]
\multicolumn{10}{l}{\emph{THRIFTY (ours)}}\\
\quad \textbf{THRIFTY}     & 0.177$_{\pm0.002}$ & \textbf{0.135}$_{\pm0.004}$ & \textbf{0.103}$_{\pm0.004}$ & \textbf{0.747}$_{\pm0.006}$ & \textbf{0.449}$_{\pm0.014}$ & \textbf{0.270}$_{\pm0.015}$ & \textbf{0.633}$_{\pm0.011}$ & \textbf{0.424}$_{\pm0.019}$ & \textbf{0.184}$_{\pm0.013}$ \\
\bottomrule
\end{tabular}}
\end{table}

\begin{table}[t]
\centering
\footnotesize
\setlength{\tabcolsep}{4pt}
\caption{\textbf{All three metrics across budget regimes}, for the standard recipe (Uniform), the
closest prior bandit (Entropy, \citealp{zuo2025strategic}), and THRIFTY: coverage@1 (problems
solved $\ge1\times$), distinct correct trajectories ($\#$dist $=\sum_q v(q)$), and per-token
efficiency (dv/1k). Cells are mean$_{\pm\text{std}}$ over $3$ seeds; dv/1k best per cell
(among these three) in bold. Table~\ref{tab:main} gives dv/1k for all five allocators.}
\label{tab:allmetrics}
\resizebox{\textwidth}{!}{%
\begin{tabular}{l ccc ccc ccc}
\toprule
& \multicolumn{3}{c}{AIME} & \multicolumn{3}{c}{MATH-500} & \multicolumn{3}{c}{LiveCodeBench}\\
\cmidrule(lr){2-4}\cmidrule(lr){5-7}\cmidrule(lr){8-10}
& cov@1 & \#dist & dv/1k & cov@1 & \#dist & dv/1k & cov@1 & \#dist & dv/1k\\
\midrule
\multicolumn{10}{l}{\emph{Scarce (25\%)}}\\
\quad Uniform              & 120$_{\pm4}$ & 168$_{\pm6}$ & \textbf{0.179}$_{\pm0.004}$ & 438$_{\pm7}$ & 671$_{\pm4}$ & 0.676$_{\pm0.001}$ & 74$_{\pm6}$ & 128$_{\pm6}$ & 0.535$_{\pm0.029}$ \\
\quad Entropy              & 113$_{\pm1}$ & 162$_{\pm5}$ & 0.172$_{\pm0.002}$ & 434$_{\pm2}$ & 665$_{\pm8}$ & 0.671$_{\pm0.007}$ & 76$_{\pm3}$ & 128$_{\pm0}$ & 0.547$_{\pm0.000}$ \\
\quad \textbf{THRIFTY}     & 120$_{\pm0}$ & 141$_{\pm4}$ & 0.177$_{\pm0.002}$ & 440$_{\pm4}$ & 649$_{\pm0}$ & \textbf{0.747}$_{\pm0.006}$ & 73$_{\pm0}$ & 146$_{\pm1}$ & \textbf{0.633}$_{\pm0.011}$ \\
\addlinespace[2pt]
\multicolumn{10}{l}{\emph{Medium (50\%)}}\\
\quad Uniform              & 140$_{\pm4}$ & 260$_{\pm6}$ & 0.126$_{\pm0.002}$ & 448$_{\pm1}$ & 856$_{\pm54}$ & 0.397$_{\pm0.009}$ & 80$_{\pm6}$ & 198$_{\pm12}$ & 0.348$_{\pm0.028}$ \\
\quad Entropy              & 136$_{\pm1}$ & 228$_{\pm3}$ & 0.118$_{\pm0.002}$ & 448$_{\pm0}$ & 860$_{\pm49}$ & 0.397$_{\pm0.006}$ & 74$_{\pm1}$ & 278$_{\pm4}$ & 0.397$_{\pm0.000}$ \\
\quad \textbf{THRIFTY}     & 137$_{\pm3}$ & 246$_{\pm16}$ & \textbf{0.135}$_{\pm0.004}$ & 453$_{\pm1}$ & 874$_{\pm33}$ & \textbf{0.449}$_{\pm0.014}$ & 78$_{\pm6}$ & 228$_{\pm15}$ & \textbf{0.424}$_{\pm0.019}$ \\
\addlinespace[2pt]
\multicolumn{10}{l}{\emph{Ample (100\%)}}\\
\quad Uniform              & 150$_{\pm5}$ & 334$_{\pm10}$ & 0.089$_{\pm0.003}$ & 453$_{\pm1}$ & 1004$_{\pm60}$ & 0.233$_{\pm0.008}$ & 81$_{\pm6}$ & 172$_{\pm13}$ & 0.160$_{\pm0.021}$ \\
\quad Entropy              & 148$_{\pm2}$ & 281$_{\pm6}$ & 0.081$_{\pm0.003}$ & 455$_{\pm1}$ & 1016$_{\pm52}$ & 0.235$_{\pm0.010}$ & 76$_{\pm2}$ & 154$_{\pm5}$ & 0.151$_{\pm0.009}$ \\
\quad \textbf{THRIFTY}     & 146$_{\pm4}$ & 330$_{\pm18}$ & \textbf{0.103}$_{\pm0.004}$ & 460$_{\pm2}$ & 1054$_{\pm38}$ & \textbf{0.270}$_{\pm0.015}$ & 84$_{\pm5}$ & 161$_{\pm12}$ & \textbf{0.184}$_{\pm0.013}$ \\
\bottomrule
\end{tabular}}
\end{table}

\FloatBarrier
\section{Conclusion}

We framed cost-constrained reasoning-data synthesis as a bandits-with-knapsacks problem and
introduced THRIFTY, a two-stage controller that spends a fixed generation-token budget where it
buys the most distinct correct solutions. A short calibration stage replaces hand-tuned knobs---the
difficulty prior, generation-length cap, and diversity threshold---with data-driven estimates, and
a diversity-gated concave objective couples allocation to an opportunity-cost retirement rule that
abandons unsolvable problems at the budget's shadow price. Across math (AIME, MATH-500) and code
(LiveCodeBench) benchmarks with a Qwen2.5-32B teacher, THRIFTY attains the best per-token
efficiency in eight of nine cells---up to $22\%$ over uniform sampling and $7$--$16\%$ over the
strongest adaptive baseline---without sacrificing coverage, and the edge persists from the scarce
into the ample-budget regime.

Our study has limitations that point to clear next steps. Most importantly, the per-query Beta
posterior THRIFTY produces measures the \emph{teacher's} success probability on each problem: it is
a difficulty signal relative to the generating model and only loosely proxies a problem's intrinsic
difficulty. It is therefore not a direct substitute for a \emph{student}-perceived difficulty when
used to order a downstream curriculum, and confirming its usefulness for student SFT requires
further experiments, which are in progress. We also evaluate at a limited set of operating points;
extending THRIFTY to larger token budgets and to teacher and student models of multiple sizes is
the natural next step toward understanding how these efficiency gains scale.

\bibliography{colm2026_conference}

@inproceedings{zelikman2022star,
 author = {Zelikman, Eric and Wu, Yuhuai and Mu, Jesse and Goodman, Noah D.},
 title = {{STaR}: Bootstrapping Reasoning with Reasoning},
 booktitle = {Advances in Neural Information Processing Systems (NeurIPS)},
 year = {2022}
}

@article{yuan2023scaling,
 author = {Yuan, Zheng and Yuan, Hongyi and Li, Chengpeng and Dong, Guanting and Lu, Keming and Tan, Chuanqi and Zhou, Chang and Zhou, Jingren},
 title = {Scaling Relationship on Learning Mathematical Reasoning with Large Language Models},
 journal = {arXiv preprint arXiv:2308.01825},
 year = {2023}
}

@article{badanidiyuru2018bandits,
 author = {Badanidiyuru, Ashwinkumar and Kleinberg, Robert and Slivkins, Aleksandrs},
 title = {Bandits with Knapsacks},
 journal = {Journal of the ACM},
 volume = {65},
 number = {3},
 year = {2018}
}

@article{zuo2025strategic,
 author = {Zuo, Bowen and Zhu, Yinglun},
 title = {Strategic Scaling of Test-Time Compute: A Bandit Learning Approach},
 journal = {arXiv preprint arXiv:2506.12721},
 year = {2025}
}

@article{gulcehre2023reinforced,
 author = {Gulcehre, Caglar and others},
 title = {Reinforced Self-Training ({ReST}) for Language Modeling},
 journal = {arXiv preprint arXiv:2308.08998},
 year = {2023}
}

@article{singh2024beyond,
 author = {Singh, Avi and others},
 title = {Beyond Human Data: Scaling Self-Training for Problem-Solving with Language Models},
 journal = {Transactions on Machine Learning Research (TMLR)},
 year = {2024}
}

@article{dong2023raft,
 author = {Dong, Hanze and others},
 title = {{RAFT}: Reward rAnked FineTuning for Generative Foundation Model Alignment},
 journal = {Transactions on Machine Learning Research (TMLR)},
 year = {2023}
}

@article{bansal2024smaller,
 author = {Bansal, Hritik and others},
 title = {Smaller, Weaker, Yet Better: Training {LLM} Reasoners via Compute-Optimal Sampling},
 journal = {arXiv preprint arXiv:2408.16737},
 year = {2024}
}

@article{snell2024scaling,
 author = {Snell, Charlie and others},
 title = {Scaling {LLM} Test-Time Compute Optimally Can Be More Effective Than Scaling Model Parameters},
 journal = {arXiv preprint arXiv:2408.03314},
 year = {2024}
}

@article{damani2025learning,
 author = {Damani, Mehul and others},
 title = {Learning How Hard to Think: Input-Adaptive Allocation of {LM} Computation},
 journal = {International Conference on Learning Representations (ICLR)},
 year = {2025}
}

@article{brown2024large,
 author = {Brown, Bradley and others},
 title = {Large Language Monkeys: Scaling Inference Compute with Repeated Sampling},
 journal = {arXiv preprint arXiv:2407.21787},
 year = {2024}
}

@inproceedings{wang2023selfconsistency,
 author = {Wang, Xuezhi and others},
 title = {Self-Consistency Improves Chain of Thought Reasoning in Language Models},
 booktitle = {International Conference on Learning Representations (ICLR)},
 year = {2023}
}

@article{matiisen2020teacher,
 author = {Matiisen, Tambet and Oliver, Avital and Cohen, Taco and Schulman, John},
 title = {Teacher-Student Curriculum Learning},
 journal = {IEEE Transactions on Neural Networks and Learning Systems},
 year = {2020}
}

@inproceedings{graves2017automated,
 author = {Graves, Alex and Bellemare, Marc G. and Menick, Jacob and Munos, R{\'e}mi and Kavukcuoglu, Koray},
 title = {Automated Curriculum Learning for Neural Networks},
 booktitle = {International Conference on Machine Learning (ICML)},
 year = {2017}
}

@inproceedings{jiang2021prioritized,
 author = {Jiang, Minqi and Grefenstette, Edward and Rockt{\"a}schel, Tim},
 title = {Prioritized Level Replay},
 booktitle = {International Conference on Machine Learning (ICML)},
 year = {2021}
}

@article{bae2025online,
 author = {Bae, Sanghwan and others},
 title = {Online Difficulty Filtering for Reasoning Oriented Reinforcement Learning},
 journal = {arXiv preprint arXiv:2504.03380},
 year = {2025}
}

@article{yu2025dapo,
 author = {Yu, Qiying and others},
 title = {{DAPO}: An Open-Source {LLM} Reinforcement Learning System at Scale},
 journal = {arXiv preprint arXiv:2503.14476},
 year = {2025}
}

@article{deepseekai2025r1,
 author = {{DeepSeek-AI}},
 title = {{DeepSeek-R1}: Incentivizing Reasoning Capability in {LLMs} via Reinforcement Learning},
 journal = {arXiv preprint arXiv:2501.12948},
 year = {2025}
}

@inproceedings{zhu2018texygen,
 author = {Zhu, Yaoming and Lu, Sidi and Zheng, Lei and Guo, Jiaxian and Zhang, Weinan and Wang, Jun and Yu, Yong},
 title = {Texygen: A Benchmarking Platform for Text Generation Models},
 booktitle = {ACM SIGIR Conference on Research and Development in Information Retrieval},
 year = {2018}
}

@inproceedings{li2016diversity,
 author = {Li, Jiwei and Galley, Michel and Brockett, Chris and Gao, Jianfeng and Dolan, Bill},
 title = {A Diversity-Promoting Objective Function for Neural Conversation Models},
 booktitle = {NAACL-HLT},
 year = {2016}
}

@article{kulesza2012determinantal,
 author = {Kulesza, Alex and Taskar, Ben},
 title = {Determinantal Point Processes for Machine Learning},
 journal = {Foundations and Trends in Machine Learning},
 volume = {5},
 number = {2--3},
 year = {2012}
}

@article{abbas2023semdedup,
 author = {Abbas, Amro and Tirumala, Kushal and Simig, D{\'a}niel and Ganguli, Surya and Morcos, Ari S.},
 title = {{SemDeDup}: Data-Efficient Learning at Web-Scale through Semantic Deduplication},
 journal = {arXiv preprint arXiv:2303.09540},
 year = {2023}
}

@inproceedings{bukharin2024data,
 author = {Bukharin, Alexander and Zhao, Tuo},
 title = {Data Diversity Matters for Robust Instruction Tuning},
 booktitle = {Findings of the Association for Computational Linguistics: EMNLP},
 year = {2024}
}

@article{thompson1933likelihood,
 author = {Thompson, William R.},
 title = {On the Likelihood that One Unknown Probability Exceeds Another in View of the Evidence of Two Samples},
 journal = {Biometrika},
 volume = {25},
 number = {3--4},
 year = {1933}
}

@article{auer2002finite,
 author = {Auer, Peter and Cesa-Bianchi, Nicol{\`o} and Fischer, Paul},
 title = {Finite-Time Analysis of the Multiarmed Bandit Problem},
 journal = {Machine Learning},
 volume = {47},
 year = {2002}
}

@article{evendar2006action,
 author = {Even-Dar, Eyal and Mannor, Shie and Mansour, Yishay},
 title = {Action Elimination and Stopping Conditions for the Multi-Armed Bandit and Reinforcement Learning Problems},
 journal = {Journal of Machine Learning Research},
 volume = {7},
 year = {2006}
}

@inproceedings{ash2020deep,
 author = {Ash, Jordan T. and Zhang, Chicheng and Krishnamurthy, Akshay and Langford, John and Agarwal, Alekh},
 title = {Deep Batch Active Learning by Diverse, Uncertain Gradient Lower Bounds},
 booktitle = {International Conference on Learning Representations (ICLR)},
 year = {2020}
}

@inproceedings{lightman2024lets,
 author = {Lightman, Hunter and Kosaraju, Vineet and Burda, Yura and Edwards, Harri and Baker, Bowen and Lee, Teddy and Leike, Jan and Schulman, John and Sutskever, Ilya and Cobbe, Karl},
 title = {Let's Verify Step by Step},
 booktitle = {International Conference on Learning Representations (ICLR)},
 year = {2024}
}

@article{jain2024livecodebench,
 author = {Jain, Naman and Han, King and Gu, Alex and Li, Wen-Ding and Yan, Fanjia and Zhang, Tianjun and Wang, Sida and Solar-Lezama, Armando and Sen, Koushik and Stoica, Ion},
 title = {{LiveCodeBench}: Holistic and Contamination Free Evaluation of Large Language Models for Code},
 journal = {arXiv preprint arXiv:2403.07974},
 year = {2024}
}

@article{otsu1979,
 author = {Otsu, Nobuyuki},
 title = {A Threshold Selection Method from Gray-Level Histograms},
 journal = {IEEE Transactions on Systems, Man, and Cybernetics},
 volume = {9},
 number = {1},
 pages = {62--66},
 year = {1979}
}

@inproceedings{fastmcts,
 author = {Li, Peiji and Lv, Kai and Shao, Yunfan and Ma, Yichuan and Li, Linyang and Zheng, Xiaoqing and Qiu, Xipeng and Guo, Qipeng},
 title = {{FastMCTS}: A Simple Sampling Strategy for Data Synthesis},
 booktitle = {Proceedings of the 63rd Annual Meeting of the Association for Computational Linguistics (ACL)},
 year = {2025}
}
\bibliographystyle{colm2026_conference}

\appendix

\section{Derivation of the per-token score}
\label{app:score}

Because each problem's value $V(q)=\ln(1+v(q))$ is concave, the budget-constrained optimum of
$\sum_q V(q)$ is a shadow-price policy: spend the next token wherever the current marginal value
per token is highest, and stop when the budget is gone. We never compute the shadow price; running
to budget realizes it, and the last affordable pull sits exactly at it. To obtain the per-token
score of a pull on $q$ at current distinct count $v$, note first that a pull yields a new distinct
solution only if it is both correct and novel, an event of probability
\[
P(\text{new distinct}) \;=\; \underbrace{p_q}_{\text{correct}}\cdot
\underbrace{q_{\mathrm{novel}}}_{\text{novel}\,\mid\,\text{correct}}.
\]
Banking one more distinct solution raises value by the finite difference
\[
\Delta V(v) \;=\; \ln(2+v)-\ln(1+v) \;=\; \ln\!\Big(1+\tfrac{1}{1+v}\Big) \;\approx\; \tfrac{1}{1+v},
\]
so the expected value of the pull is $p_q\,q_{\mathrm{novel}}\,\tfrac{1}{1+v}$. Dividing by the
expected token cost $c_q$ of a pull gives the marginal value per token, our selection score,
$\mathrm{score}(q)=\tilde p_q\,q_{\mathrm{novel}}(q)\,\tfrac{1}{1+v(q)}\,\tfrac{1}{c_q}$. The
solvability term is a Thompson draw $\tilde p_q\sim\mathrm{Beta}(\alpha_q,\beta_q)$
\citep{thompson1933likelihood} rather than the posterior mean, which preserves the exploration a
point estimate would lose; $q_{\mathrm{novel}}$ is estimated by the running ratio $v/n_q^{+}$ (set
to $1$ before the first correct trace); $1/(1+v)$ is the closed-form marginal of the value
function; and $c_q$ is the running mean completion tokens per pull (the global mean before a
problem's first pull). Relative to a plain Thompson score $\tilde p_q$, the change is surgical:
two multiplicative factors and a cost normalizer are added, and the active-set filter drops its
$n_q^{+}<K$ clause, keeping only $\mathrm{active}(q)=\lnot\,\mathrm{retired}(q)$.

\end{document}